\documentclass[]{spie}  

\usepackage{amsmath,amsfonts,amssymb}
\usepackage{graphicx}
\usepackage{acronym}
\usepackage[colorlinks=true, allcolors=blue]{hyperref}
\usepackage{tabularray}
\usepackage{color}
\usepackage{subfigure}

\definecolor{TreeEvergreen}{rgb}{0,0.392,0}
\definecolor{TreeDeciduous}{rgb}{0,0.627,0}
\definecolor{Shrubs}{rgb}{0.588,0.392,0}
\definecolor{Grasslands}{rgb}{1,0.705,0.196}
\definecolor{Croplands}{rgb}{1,1,0.392}
\definecolor{GrassVegetation}{rgb}{0,0.862,0.509}
\definecolor{BareAreas}{rgb}{1,0.96,0.843}
\definecolor{Builtup}{rgb}{0.764,0.078,0}
\definecolor{WaterSeasonal}{rgb}{0.356,0.584,1}
\definecolor{WaterPermanent}{rgb}{0,0.274,0.784}
\definecolor{Alto}{rgb}{0.93,0.93,0.93}
\newcommand{\cred}[1]{\textcolor{black}{#1}}
\newcommand{\cblue}[1]{\textcolor{black}{#1}}

\acrodef{ResNet}{Residual Neural Network}
\acrodef{RS}{Remote Sensing}
\acrodef{TS}{Time Series}
\acrodef{CCE}{Categorical Cross-Entropy}
\acrodef{NLL}{Negative Log-Likelihood}
\acrodef{FT}{Fine-Tuning}
\acrodef{CCI}{Climate Change Initiative}
\acrodef{HRLC}{High Resolution Land Cover}
\acrodef{TPE}{Tree-structured Parzen Estimator}
\acrodef{UA}{User's Accuracy}
\acrodef{PA}{Producer's Accuracy}
\acrodef{F1}{F1-Score}
\acrodef{VHR}{Very High-Resolution}
\acrodef{HR}{High-Resolution}

\title{A class-driven hierarchical ResNet for classification of multispectral remote sensing images}

\author{Giulio Weikmann}
\author{Gianmarco Perantoni}
\author{Lorenzo Bruzzone}
\affil{Department of information engineering and computer science, University of Trento, Via Sommarive, 9 I-38123, Trento, Italy}

\authorinfo{Proc. SPIE 12733, Image Sign. Proc. Remote Sens. XXIX, 127330D, Preprint, Full version: \href{https://doi.org/10.1117/12.2679293}{10.1117/12.2679293}.}

\begin{document} 
\maketitle

\begin{abstract}
This work presents a multitemporal class-driven hierarchical \ac{ResNet} designed for modelling the classification of \ac{TS} of multispectral images at different semantical class levels. The architecture consists of a modification of the \ac{ResNet} where we introduce additional branches to perform the classification at the different hierarchy levels and leverage on hierarchy-penalty maps to discourage incoherent hierarchical transitions within the classification. In this way, we improve the discrimination capabilities of classes at different levels of \cred{semantic} details and train a modular architecture that can be used as a backbone network for \cred{introducing} new \cred{specific} classes and additional tasks \cred{considering limited training samples available}. We exploit the class-hierarchy labels to train efficiently the different layers of the architecture, allowing the first layers to train faster on the first levels of the hierarchy \cred{modeling general classes} (i.e., the macro-classes) \cred{and the intermediate classes}, while using the last ones to discriminate \cred{more specific} classes (i.e., the micro-classes). \cred{In this way, the targets are constrained in following the hierarchy defined, improving the classification of classes at the most detailed level.} The proposed \cred{modular} network \cred{has intrinsic} adaptation capability \cred{that can be obtained through fine tuning}. The experimental results, obtained on two tiles of the Amazonian Forest on 12 monthly composites of Sentinel 2 images acquired during 2019, demonstrate the effectiveness of the hierarchical approach in both generalizing over different hierarchical levels and learning discriminant features for an accurate classification at the micro-class level \cred{on a new target area, with a better representation of the minoritarian classes}. 
\end{abstract}


\section{INTRODUCTION}
\label{sec:intro}  
In the \ac{RS} field, deep learning has been proved to be a very successful methodology to deal with EO applications, having the capabilities to solve highly complex classification tasks and removing the need of defining hand-crafted features. Convolutional Neural Networks have shown promising results in the extraction of both mid and high-level abstract features in image recognition \cite{8270691,BOULILA2021106014}, object detection \cite{DENG20183,7560644}, and semantic segmentation \cite{9122009}. Such architectures allow the combination of lower and higher level of semantic features, by exploiting the local connectivity of the pixels in the images. In this context, CNNs have the intrinsic ability of learning features hierarchically \cite{bilal_jourabloo_ye_liu_ren_2018}, allowing the definition of different levels of classification. However, one of the drawbacks of training a network from scratch is \cred{the requirement to have a} huge number of labelled samples, which should be proportional to the number of parameters of the architecture \cite{caruana2000overfitting}. \cred{This} is not always feasible in the tasks considered. \cred{In particular, in \ac{RS}} the total amount of available training data is far less than that of other fields due to the lack of dataset annotation, which require high interpretation ability and expert knowledge. As a consequence, several studies derive architectures from other fields and adapt them to the new tasks \cite{9782149,farahbakhsh2020computer}. Common approaches derive pre-trained models from the computer vision field \cite{bazi2019simple}, applying few-shot learning techniques to learn from a small number of labelled samples through data augmentation and knowledge re-use \cite{9328476}. One common approach to leverage knowledge gained from models trained on different fields consists in the fine-tuning of a pre-trained architecture \cite{liu2018classifying}. Here, the weights of the network are frozen, and the classification layers are substituted to take into account the different target classes. The network is then progressively unfrozen and is re-trained using a small training set and learning rate. By combining knowledge re-use and data augmentation, effective \ac{RS} models can be developed even with limited labelled data. \cred{Moreover,} most of the models fine-tuned in the \ac{RS} field adapt the pre-trained model for object detection and scene recognition \cred{in natural images} considering \ac{VHR} images \cite{zhang2023generalized, CHEN2023105549}, which focus on \cred{three} visible spectral channels \cred{(Red, Green, Blue)} and have better geometric resolutions compared to \ac{HR} multispectral images. The adoption of pre-trained architectures tailored to natural images leads to suboptimal results due to the peculiar characteristics of \ac{HR} \ac{RS} data: (i) natural images and \ac{RS} data represent information at different geometric resolutions, (ii) multispectral \ac{RS} images contain much more spectral information than natural images, and (iii) \ac{RS} data are geolocated. \cred{In the \ac{RS}, considerable efforts \cite{tuia2016domain} have been dedicated in adapting \ac{RS} tailored architectures to new spatial images and tasks. Such methods rely on the selection of invariant features during the training, the adaptation of data distributions, and the adaptation of the classifiers. Despite demonstrating good performance, the exploration of this field is still ongoing, primarily owing to the critical role of data adaptation. The importance of refining data adaptation techniques to extract invariant features suggests that further investigation is required.} In this context, the development of a general architecture tailored to the specific characteristics of multispectral data being able to perform hierarchical feature extraction and classification would facilitate the training on new target tasks by reducing the number of labelled samples required to adapt the network. 

Residual neural networks are well-suited for training with small datasets due to their residual connections, which partially alleviates the challenges of training deep neural networks with limited data. The use of skip connections mitigates the problem of vanishing gradient in deep neural networks and allows for an easier optimization during training, reducing the risk of overfitting small datasets. Moreover, such ability to learn and optimize on small datasets leads to better generalization \cite{8984747} and parameter efficiency \cite{allen2019can}. \cblue{These} \cred{architectures are suitable for the hierarchical extraction of features from images, since their learning is similar to that of simpler CNNs. \cblue{Indeed}, different works aim at extracting features hierarchically at different levels \cite{zhu2017b,SEO2019328}, leveraging the concept that lower layers in the architectures tend to learn more generic features such as edges, textures and basic shapes, which are often relevant to a wide range of tasks \cite{wang2020cnn}. Though such methods showed good performance, they are \cblue{mainly} tailored to natural images. It is worth noting that the hierarchical learning of the features is performed independently at each hierarchical stadium, detached from the concurrent classification of other strata.}

In this work, we aim to exploit the intrinsic hierarchy of land-cover classes to guide the learning of discriminant features at different depths of the architecture. The main idea is to distribute the learning of the features at different levels within the architecture, exploiting the hierarchy to improve the discrimination capabilities between macro-classes and micro-classes. In this way, we impose the hierarchical extraction of features, driving the network to learn more general features in the early layers to discriminate macro-classes, while extracting more specific features for discriminating micro-classes as the depth of the network increases. The final architecture is modularly trained, allowing the user to detach blocks of the network and use the macro-layers and intermediate-layers as backbone for the optimization of the network on new applications and micro-classes. The proposed network also combines hierarchy-penalty terms to reward a hierarchy consistent classification\cite{perantoni2021novel}.

The structure of the rest of the paper is as follows. Section \ref{sec:proposed_approach} presents the proposed system architecture for land cover classification. Section \ref{sec:dataset} describes the considered study area for the training of the network and the fine-tuning. Section \ref{sec:setup} presents the experimental setup defined to test the effectiveness of the proposed approach, while the experimental results obtained are discussed in Section \ref{sec:results}. Finally, Section \ref{sec:conclusion} reports the finding and concludes the paper. 

\section{Methodology}
\label{sec:proposed_approach}

\begin{figure}{
\centering
    \includegraphics{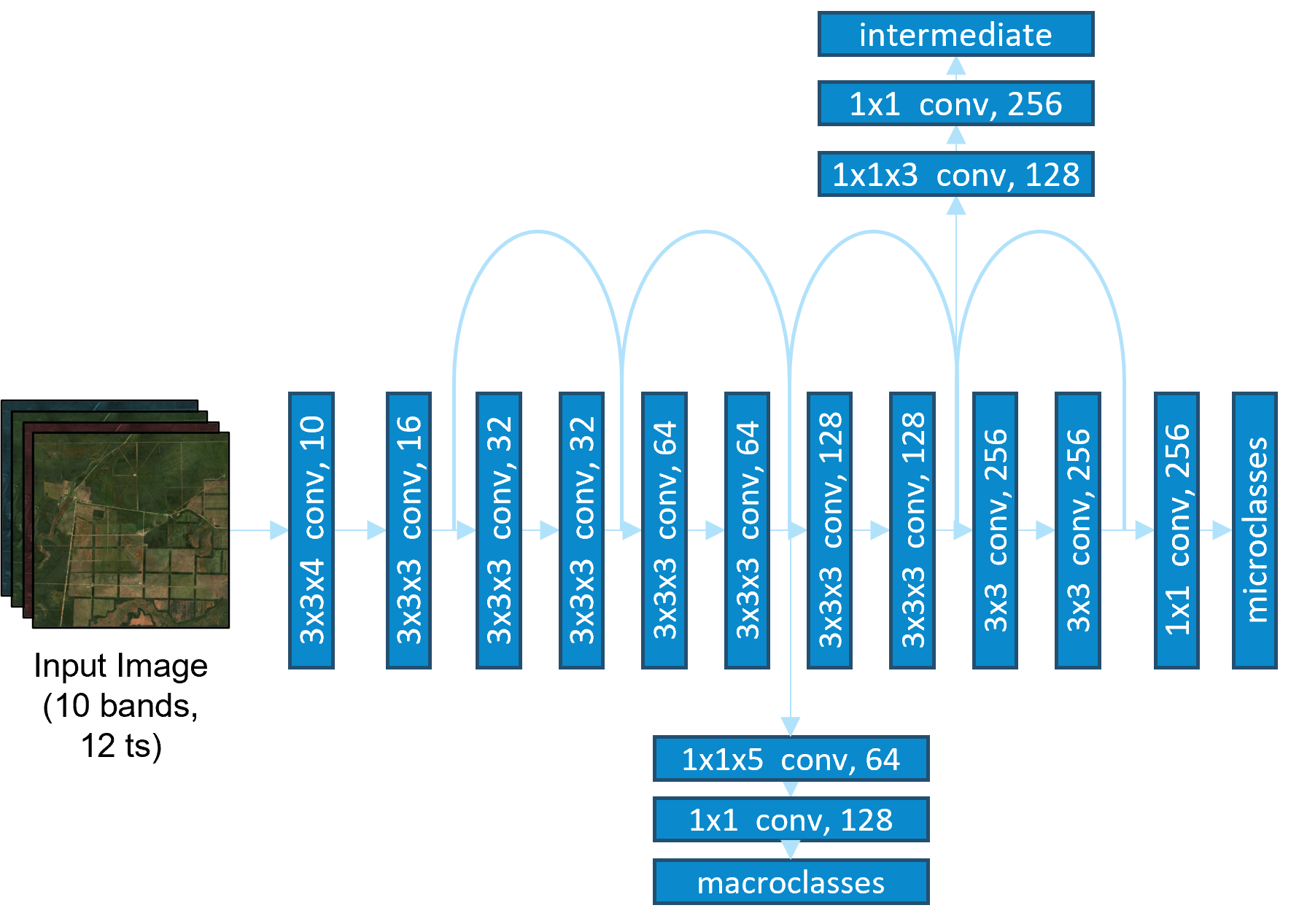}
    \caption{Architecture of the proposed Class-driven Hierarchical \ac{ResNet}.
    }\label{fig:ResNet}}
\end{figure}

The proposed approach is based on a temporal \ac{ResNet} \cred{that is modified} to take into account the semantical hierarchy definition of the class targets. Here, we propose a temporal class-driven Hierarchical \ac{ResNet} for the classification of multispectral images at different levels of detail. The architecture, which is shown in Fig. \ref{fig:ResNet}, consists of a modified \ac{ResNet} where we introduce additional branches to perform the classification at different levels of the classes organized hierarchically. The classes considered are aggregated in a semantically hierarchical way, where the less detailed level is related to the macro-classes, which can be further subdivided into micro-classes at a more specific level of details. \cred{For illustration purposes, let \cblue{us} consider three target micro-classes representing: ``asphalt \cblue{roads}", ``\cblue{buildings with red roof}", and ``\cblue{buildings with grey roof}". These classes can be aggregated in ``\cblue{built-ups}" and ``\cblue{roads}" at the intermediate level, and furtherly aggregated in the ``\cblue{non-vegetated areas}" macro-class.} We exploit such semantically aggregated hierarchy labels to train efficiently the different layers of the architecture, allowing the first layers to train faster on the first levels of the hierarchy (i.e., the macro-classes), while using the last ones to discriminate the individual classes (i.e., the micro-classes). To this end, the proposed network combines hierarchy-penalty terms and hierarchy classifications to reward a hierarchy-consistent feature selection. The penalty term, which is added to the loss function, consists in the cross entropy calculated between the estimated class posterior probabilities obtained at the micro-class layer re-projected onto the precedent hierarchy level through a penalty matrix that maps the correct transitions. 

\subsection{Class-driven Hierarchical ResNet}
In the \cred{proposed} system architecture, we adopt a multitemporal hierarchical class-driven approach using a \ac{ResNet}. \cred{Although the presented methodology is suitable for individual acquisitions, we specifically showcase its application in the context of \ac{TS} analysis using monthly composites of Sentinel-2 images.} This choice allows us to effectively capture the intricate temporal patterns present in the dataset and enables hierarchical classification of various temporal land-cover classes. The selection of \ac{ResNet} is motivated by its ability to accurately learn from limited data and its capability to generalize well across different scenarios. The class-driven \ac{ResNet} takes as input Sentinel-2 patches of $30 \times 30$ pixels, representing a \ac{TS} of 12 monthly composites with 10 spectral channels. The network consists of a 3D convolutional layer, followed by a \ac{ResNet}8 and a 1D convolutional layer to extract the features at the micro-class layer. In this version of the \ac{ResNet}, we exclude any pooling operation, in order to prevent overfitting \cite{he2016deep}. Additionally, we introduce temporal convolutions in the skip connection to extract the temporal features from the \ac{TS}. The \ac{ResNet} progressively compacts the temporal dimensions by not performing padding operations on the temporal channels. Additional branches are introduced to perform the classification at the intermediate and macro-class levels. These layers are composed of a temporal convolution to downsample the time features and are followed by 1D convolution for the classification of the feature extracted. 
To train the network, we considered three standard \ac{CCE} losses (one for each classification layer: \cred{the micro-class layer, intermediate-class layer, and the macro-class layer specifically}) with a penalty term to correctly map the hierarchy consistent transitions.
\subsection{Hierarchy Penalty Term}
\label{sec:penalty}
To obtain a hierarchy consistent classification at the three different levels, an approach relying on additional penalty terms has been selected. \cred{The penalty terms are defined by} transition matrices defined to map the correct transition from the different layers. Let us consider \cred{for illustration purposes} the case of two target hierarchy levels, one representing the macro classes and one the micro classes. 

Let $M$ be the set of $k$ macro-class labels, where $M = \{M_1, M_2, \dots, M_k\}$, and $m$ be the set of $n$ micro-class labels, where $m = \{m_1, m_2, \dots, m_n\}$. We can define the transition matrix $T$ of size $n \times k$, where $T_{i,j}$ denotes the probability of transition from micro-class $m_i$ to macro-class $M_j$. The transition matrix $T$ can be represented as:
\[
T =
\begin{bmatrix}
T_{1,1} & T_{1,2} & \dots & T_{1,k} \\
T_{2,1} & T_{2,2} & \dots & T_{2,k} \\
\vdots & \vdots & \ddots & \vdots \\
T_{n,1} & T_{n,2} & \dots & T_{n,k} \\
\end{bmatrix}
\]

The transition matrix $T$ can be defined as the conditional probability of transition \cred{given} a micro-class $m_i$ to a macro-class $M_j$ \cred{as is} denoted as $P(M_j \mid m_i)$, i.e. $T_{i,j} = P(M_{j} \mid m_{i})$. The probability distribution over the macro-classes can be obtained from the probability distribution over the micro-classes and the transition matrix through the following:
\begin{equation}
    P(M_j \mid x) = \sum_{i=1}^{n} P(M_j \mid m_i) \cdot P(m_i \mid x)
\end{equation}
where \cred{$P(M_j \mid x)$} is the probability that a sample $x$ belongs to the macro-class $M_j$, and it is obtained by summing over all micro-classes $m_i$ considering their probabilities \cred{$P(m_i \mid x)$} and the transition probabilities \cred{$P(M_j \mid m_i)$}.
The transition matrices have been defined for each couple of hierarchy \cred{classes (i.e., micro-classes to intermediate, intermediate to macro-classes and micro-classes to macro-classes), modeling} both the forward projection of the logits (from macro-class to micro-class) and the backward projection of the logits (from micro to macro) considering transposed matrices. We map the permitted transitions with a probability of $1$, \cred{whereas} the wrong transitions are mapped with zero probability. The transition matrices \cred{are} then normalized by row, ensuring that $\sum_{j=1}^n T_{i,j}=1$.

To reproject the logits retrieved by the architecture on the different hierarchical level of classification, we exploit the \ac{NLL} loss between the target class and the reprojected logits. Let $z_j$ be the logits calculated at the micro-class classification layer. We calculated the logits reprojection on the macro-class layer $\dot z_i$ as:
\begin{equation}
    \dot z_i = \log{\sum_j{\exp{(z_j+\log{T_{i,j}}})}} - \log{\sum_j{\exp{z_{j}}}}
\end{equation}
which can then be used to calculate the \ac{NLL} loss \cred{as follows}:
\begin{equation}
\text{NLLLoss}(\dot z_i, \mathbf{t}) = -\log\left(\frac{\exp(\mathbf{\dot{z_i}}_t)}{\sum_{c=1}^{C}\exp(\mathbf{\dot{z_i}}_{c})}\right)
\end{equation}
where $\mathbf{\dot{z_i}} \in \mathbb{R}^{C}$ is the predicted log-probability distribution (logits) for each class, $C$ is the number of classes at the reprojected layer, and $\mathbf{t} \in \mathbb{N}$ is the target class label.

The \ac{NLL} losses calculated at the three hierarchical levels defined in the architecture are then added as a penalty term to the cumulative loss of the network. To further improve the learning of features at the different hierarchical levels, both the \ac{CCE} and the \ac{NLL} are multiplied by a weighting factor optimized during the training.  

\subsection{Modularity and Fine-Tuning Strategy}
\label{sec:finetuning}
\cred{The intrinsic modularity nature of the proposed architecture allows for a refinement of the model at the hierarchical semantic level needed, aligning seamlessly with the user's specific classification requirements. This adaptability lets the network adjust to new classes and objectives while still following the original hierarchy \cblue{and parameter} set during training, with only minor changes needed.}

\cred{In this context, we employ a \ac{FT} approach, improving the network's performance on \cblue{either} fresh targets or domain shifts, all while working with a restricted set of training samples. The hierarchical ResNet architecture enables model pruning at the hierarchical classification level corresponding to the user's application demands. Through this strategy and by exploiting the capability of the network to hierarchically assimilate features, the proposed architecture can accommodate various levels of abstraction. This allows the flexibility to select the optimal level of detail for the given classification task, while maintaining the knowledge gained during the training on both minoritarian and majoritarian classes without the requirement of training a new network from scratch.}

The fine-tuning of the network is carried out in three steps:
\begin{itemize}
\item \cblue{\textit{Step 1 --- Pruning the hierarchical level under analysis.}} \cred{This consists in the removal of the classification layer tailored to the classification at the hierarchical level required, allowing the incorporation of new classes. Subsequently, a new 1D convolutional layer with the output nodes corresponding to the number of classes in the new classification scheme \cblue{is instantiated}.}
\item \cblue{\textit{Step 2 --- Freeze the network weights to prevent significant deviations from the original feature space learned by the network.}} This ensures that the existing knowledge is retained, as well as the hierarchical representation of the problem at hand. \cred{During this step, an initial training of the network is performed, to integrate the new classes with the previously learned features. This mitigates the risk of overfitting and catastrophic forgetting, as well as allowing feature reusability.}
\item \cblue{\textit{Step 3 --- Unfreeze the entire network and perform the learning}}. This allows for further fine-tuning across all layers, maximizing the adaptation capability to the new target or domain.
\end{itemize}

\section{Dataset Description}
\label{sec:dataset}

\begin{table}
\caption{\cred{Considered} classification scheme and distribution of the samples for the micro-class hierarchy level for the two selected tiles. The macro-class aggregation is represented by the borders grouping.}
\label{tab:classification-scheme}
\begin{tblr}{
  column{3} = {c},
  column{4} = {c},
  column{5} = {c},
  column{6} = {c},
  column{7} = {c},
  column{8} = {c},
  cell{1}{3} = {c=3}{},
  cell{1}{6} = {c=3}{},
  cell{2}{3} = {c=2}{},
  cell{2}{6} = {c=2}{},
  cell{3}{1} = {r},
  cell{3}{2} = {TreeEvergreen},
  cell{4}{1} = {r},
  cell{4}{2} = {TreeDeciduous},
  cell{5}{1} = {r},
  cell{5}{2} = {Shrubs},
  cell{6}{1} = {r},
  cell{6}{2} = {Grasslands},
  cell{7}{1} = {r},
  cell{7}{2} = {Croplands},
  cell{8}{1} = {r},
  cell{8}{2} = {GrassVegetation},
  cell{9}{1} = {r},
  cell{9}{2} = {BareAreas},
  cell{10}{1} = {r},
  cell{10}{2} = {Builtup},
  cell{11}{1} = {r},
  cell{11}{2} = {WaterSeasonal},
  cell{12}{1} = {r},
  cell{12}{2} = {WaterPermanent},
  vline{4} = {1}{},
  vline{6} = {2}{},
  vline{2-3,6,9} = {3-12}{},
  hline{2} = {3-8}{},
  hline{3,6,9,11,13} = {2-8}{},
  hline{4-5,7-8,10,12} = {2}{},
}
 &  & \textbf{21KUQ} &  &  & \textbf{21KXT} &  & \\
 &  & HRLC &  & GT & HRLC &  & GT\\
\textbf{Tree cover evergreen} & \hphantom{-} & 24673132 & 23.49\% & 390 & 13774780 & 13.12\% & 256\\
\textbf{Tree cover deciduous} &  & 29418864 & 28.01\% & 335 & 16767839 & 15.97\% & 249\\
\textbf{Shrub cover} &  & - & - & - & 2146001 & 2.04\% & 73\\
\textbf{Grasslands} &  & 45197572 & 43.04\% & 521 & 68232899 & 64.97\% & 717\\
\textbf{Croplands} &  & 4937739 & 4.70\% & 127 & 3013630 & 2.87\% & 111\\
\textbf{Grass. veg. aq. or flooded} &  & 388438 & 0.37\% & 360 & 471607 & 0.45\% & 152\\
\textbf{Bare areas} &  & 194850 & 0.19\% & 91 & 140142 & 0.13\% & 42\\
\textbf{Built-up} &  & 22999 & 0.02\% & 51 & 225142 & 0.21\% & 158\\
\textbf{Open water seasonal} &  & 19298 & 0.02\% & 43 & 63926 & 0.06\% & 73\\
\textbf{Open water permanent} &  & 168612 & 0.16\% & 237 & 185538 & 0.18\% & 351
\end{tblr}
\end{table}

The study area focuses on two specific tiles acquired in the Amazonian region. In particular, the remote sensing data utilized for the study were \ac{TS}s of Sentinel 2 images acquired over two tiles, namely ``21KUQ" and ``21KXT". The separation between the two tiles is $450km$, measured from the centre of each tile, which guarantees a lack of correlation among the analysed samples. Furthermore, the two examined scenes are situated in distinct ecoregions characterized by varying climatic conditions. To train the network, a total of twelve monthly composites were generated, covering the period from January to December 2019. The composites were created using 10 spectral channels at $10m$ and $20m$, with the latter resampled to match the $10m$ resolution. 

The ground truth considered consists in a few photointerpreted reference pixels over the two tiles adopted, while the reference land cover maps adopted were generated in the \ac{CCI} \ac{HRLC} project\footnote{\href{https://climate.esa.int/en/projects/high-resolution-land-cover/}{ESA Climate Office, High Resolution Land Cover}} at $10m$ of resolution \cite{bruzzone2022esa}. The land-cover maps were generated considering a fusion of Sentinel-2 optical imagery and Sentinel-1 \cred{SAR} data. The dataset has been validated through independent data sources. Nonetheless, the photointerpreted ground truth exhibits limited dispersion across the two examined tiles, with numerous samples obtained from comparable regions within the analysed images. As a result of this limitation, the ground truth may \cred{lack in precision} to adequately encompass the intricate spectral characteristics of the diverse hierarchical classes found in the scene. 

We considered two similar classification schemes for the two tiles analysed, with an additional micro-class target in the fine-tuned tile, to emphasize the \cred{adaptability} of the network. Table \ref{tab:classification-scheme} shows the micro-class classification scheme, as well as the distribution of the classes in the two tiles analysed. The considered classes show a high similarity in their spectral signatures, which requires the introduction of the temporal component to perform an accurate classification and discrimination. The micro-classes are then aggregated into five intermediate classes, which are further aggregated into 4 macro-classes, consisting of "Trees-Shrubs", "Other Vegetation", "Non-Vegetated Area", and "Water bodies". The grouping of the micro-classes in intermediate and macro-classes emerges from both a semantic analysis of the micro-classes and similarities found in the spectral characteristics. One can note that the dataset is highly imbalanced, with most of the tiles covered by the "Tree cover" and "Grasslands" classes, which extend over 94\% of each of the tiles analysed, introducing severe challenges in the problem considered. We omitted the data related to shrub cover within the ``21KUQ" tile due to the minimal presence of shrubs in that area. This decision was also made to underscore the potential for integrating new classes within the hierarchy established in the fine-tuned tile. 

\section{Design of the experiments}
\label{sec:setup}
This section presents in detail the design of the experiments during (i) the training of the architecture from scratch considering the tile ``21KUQ", and (ii) the fine-tuning of the hierarchical architecture on the tile ``21KXT" considering a limited amount of training samples available. 

\subsection{Training Phase}
For the training phase, we \cred{compared the proposed} Hierarchical Class-driven \ac{ResNet} \cred{with} a standard \ac{ResNet} without the three hierarchical classification layers. To ensure a spatial decorrelation between the training set and the test set, we subdivided the ``21KUQ" tile in patches of size $30 \times 30$, from which we removed the patches containing the photointerpreted samples. We allocated 30\% randomly selected patches for training the architecture, while the remaining 70\% for the test set. We monitored the training of the architecture using the photointerpreted samples as validation set, and an early-stopping technique was implemented to halt the training process when the loss on the validation set ceased to decrease. 

Due to the peculiar highly imbalanced problem under analysis, we adopted an oversampling approach that leverages the label entropy within the input patches. In the training set, patches showing a diverse variety of different labels were favoured, as opposed to those representing only singular classes. To enhance the oversampling of less frequent and more diverse patches, the patches were subsequently weighted based on the most present label's prior probability within each patch. 

For the optimization of the hyperparameters of both the architectures, we employed a \ac{TPE} algorithm to efficiently explore the hyperparameter search space. The learning rate and the weight decay were sampled from log-uniform distribution, $U_{log}([10^{-3},10^{-5}])$ and $U_{log}([10^{-6},10^{-2}])$ respectively. The weights for the cross-entropies and the weights associated with the penalty terms calculated in the \cred{modular} class-driven \ac{ResNet} were sampled from uniform distributions in the interval $U([1,3])$. The optimal setup was identified according to the accuracy achieved on the validation set, considering five training epochs. After the parameters have been fixed for both the standard and the proposed methodologies, we trained the architecture considering the early-stopping technique.

\subsection{Fine-Tuning}

After fixing the hyperparameters and training the models on the ``21KUQ" tile, the model obtaining the best performance on the validation set was retrieved to perform the fine-tuning on the ``21KXT" tile. In these experiments, we pruned the micro-class classification layer of the class-driven hierarchical \ac{ResNet} and substituted it with a 1D convolutional layer with an additional channel output, to account for the additional ``Shrub" micro-class. 

To showcase the adaptation capability and the modularity of the pre-trained class-driven \ac{ResNet}, we considered a small training set consisting of the photointerpreted ground truth samples. In this scenario, we fine-tune the architecture considering 2109 samples in total, which prior distribution can be seen in Table \ref{tab:classification-scheme}. As previously discussed, the selected training set still shows an imbalanced dataset, with the majority of samples belonging to majoritarian classes (``Grassland" and ``Trees"), while instead limited samples are available for minoritarian classes. \cred{Additionally, it is important to highlight that the photointerpreted samples are not uniformly spread throughout the analysed tile. Instead, multiple samples are concentrated within the same area}. This approach positions us in a scenario where we are working with a limited training set, making it impractical to train a deep architecture from scratch. 

Following the procedure described in Sec. \ref{sec:finetuning}, we froze the network's weight except for the weights associated to the new instantiated \cred{micro-classes} classification layer. We trained the micro-class classification layer for 20 iterations before unfreezing the rest of the architecture. In this phase, the intermediate and macro-class layers are ignored to leverage on the classification of the micro-class layer. Training the intermediate and macro-class layers would require more training samples, to correctly extrapolate the hierarchical features of the different classes, which instead are already learned from the pre-trained network. The entire network is then unfrozen, in \cred{an adaptive step aimed} to better fit the new classification task. By exploiting the hierarchical features obtained by the class-driven \ac{ResNet}, we leverage on the hierarchical features extracted \cred{in the first layers of the architecture, which corresponds to more generic features learned during the training, to optimize the features learned at the last classification layer}. Following the fine-tuning of the network with unfrozen layers, we tracked the weighted validation loss on the validation set. This approach was selected to address the challenge posed by the imbalanced dataset and ensure that the loss was unbiased due to the different prior of the classes. After reaching a minimum in the validation loss, we proceeded to evaluate the models on the independent test set.

For the optimization of the hyperparameters in the fine-tuning phase, the \ac{TPE} algorithm was exploited. However, in this experiment, the weight for the cross-entropies and the weights associated with the penalty terms were fixed to the best configuration found in the training step. The learning rate and weight decay were sampled similarly to the previous configuration, leveraging on log-uniform distributions. For the Std. \ac{ResNet} we employed the same approach, i.e. fine-tuning the best \ac{ResNet} obtained in the previous iteration to compare the capability of both the networks to adapt to new problems. We also compared the results retrieved by the fine-tuned networks with an architecture trained from scratch, showing that the pre-trained network obtains higher performance, fitting better the new classification task when the training is complete. 

\section{Experimental results}
\label{sec:results}
\begin{table}
\centering
\caption{Comparison of the best \ac{UA} and \ac{PA} accuracies obtained on the photointerpreted ground truth in the tile ``21KUQ" \cred{on} the three hierarchical aggregations of the classes adopted, considering the Std. \ac{ResNet} and the proposed Hierarchical Class-driven \ac{ResNet}.}
\label{tab:training}
\resizebox{\textwidth}{!}{
\begin{tblr}{
  column{3} = {c},
  column{4} = {c},
  column{5} = {c},
  column{6} = {c},
  column{7} = {c},
  column{8} = {c},
  column{9} = {c},
  column{10} = {c},
  column{11} = {c},
  column{12} = {c},
  column{13} = {c},
  column{14} = {c},
  cell{1}{3} = {c=12}{},
  cell{2}{3} = {c=6}{},
  cell{2}{9} = {c=6}{},
  cell{4}{1} = {r},
  cell{4}{2} = {TreeEvergreen},
  cell{4}{3} = {Alto},
  cell{4}{5} = {r=2}{},
  cell{4}{6} = {r=2}{Alto},
  cell{4}{7} = {r=2}{},
  cell{4}{8} = {r=2}{Alto},
  cell{4}{10} = {Alto},
  cell{4}{11} = {r=2}{Alto},
  cell{4}{12} = {r=2}{},
  cell{4}{13} = {r=2}{Alto},
  cell{4}{14} = {r=2}{},
  cell{5}{1} = {r},
  cell{5}{2} = {TreeDeciduous},
  cell{5}{4} = {Alto},
  cell{5}{9} = {Alto},
  cell{6}{1} = {r},
  cell{6}{2} = {Grasslands},
  cell{6}{3} = {Alto},
  cell{6}{5} = {r=2}{},
  cell{6}{6} = {r=2}{},
  cell{6}{7} = {r=3}{},
  cell{6}{8} = {r=3}{},
  cell{6}{10} = {Alto},
  cell{6}{11} = {r=2}{Alto},
  cell{6}{12} = {r=2}{Alto},
  cell{6}{13} = {r=3}{Alto},
  cell{6}{14} = {r=3}{Alto},
  cell{7}{1} = {r},
  cell{7}{2} = {Croplands},
  cell{7}{4} = {Alto},
  cell{7}{9} = {Alto},
  cell{8}{1} = {r},
  cell{8}{2} = {GrassVegetation},
  cell{8}{3} = {Alto},
  cell{8}{5} = {Alto},
  cell{8}{10} = {Alto},
  cell{8}{12} = {Alto},
  cell{9}{1} = {r},
  cell{9}{2} = {BareAreas},
  cell{9}{3} = {Alto},
  cell{9}{5} = {r=2}{Alto},
  cell{9}{6} = {r=2}{},
  cell{9}{7} = {r=2}{Alto},
  cell{9}{8} = {r=2}{},
  cell{9}{10} = {Alto},
  cell{9}{11} = {r=2}{},
  cell{9}{12} = {r=2}{Alto},
  cell{9}{13} = {r=2}{},
  cell{9}{14} = {r=2}{Alto},
  cell{10}{1} = {r},
  cell{10}{2} = {Builtup},
  cell{10}{3} = {Alto},
  cell{10}{10} = {Alto},
  cell{11}{1} = {r},
  cell{11}{2} = {WaterSeasonal},
  cell{11}{3} = {Alto},
  cell{11}{5} = {r=2}{},
  cell{11}{6} = {r=2}{Alto},
  cell{11}{7} = {r=2}{},
  cell{11}{8} = {r=2}{Alto},
  cell{11}{10} = {Alto},
  cell{11}{11} = {r=2}{Alto},
  cell{11}{12} = {r=2}{},
  cell{11}{13} = {r=2}{Alto},
  cell{11}{14} = {r=2}{},
  cell{12}{1} = {r},
  cell{12}{2} = {WaterPermanent},
  cell{12}{4} = {Alto},
  cell{12}{9} = {Alto},
  cell{13}{1} = {r},
  cell{13}{3} = {Alto},
  cell{13}{10} = {Alto},
  cell{13}{11} = {Alto},
  cell{13}{12} = {Alto},
  cell{13}{13} = {Alto},
  cell{13}{14} = {Alto},
  cell{14}{1} = {r},
  cell{14}{3} = {c=2}{},
  cell{14}{5} = {c=2}{},
  cell{14}{7} = {c=2}{},
  cell{14}{9} = {c=2}{Alto},
  cell{14}{11} = {c=2}{Alto},
  cell{14}{13} = {c=2}{Alto},
  vline{4,10} = {2}{},
  vline{9} = {1,2,3,13-14}{},
  vline{2-3,9} = {4,5,6,7,8,9,10,11,12}{},
  vline{2-3} = {5,7-8,10,12}{},
  hline{3,14} = {3-14}{},
  hline{4} = {2-14}{},
  hline{5-12} = {2}{},
  hline{13} = {-}{},
}
 &  & Photointerpreted GT &  &  &  &  &  &  &  &  &  &  & \\
 &  & Std. ResNet &  &  &  &  &  & Class-driven ResNet &  &  &  &  & \\
 &  & {Micro\\UA(\%)} & {Micro\\PA(\%)} & {Int.\\UA (\%)} & {Int.\\PA (\%)} & {Macro\\UA (\%)} & {Macro\\PA (\%)} & {Micro\\UA(\%)} & {Micro\\PA(\%)} & {Int.\\UA (\%)} & {Int.\\PA (\%)} & {Macro\\UA (\%)} & {Macro\\PA (\%)}\\
Tree cover  evergreen &  & \textbf{85.34} & 83.59 & 97.81 & \textbf{98.48} & 97.81 & \textbf{98.48} & 82.35 & \textbf{86.15} & \textbf{98.21} & 98.21 & \textbf{98.21} & 98.34\\
Tree cover deciduous &  & 80.75 & \textbf{83.88} &  &  &  &  & \textbf{82.59} & 77.91 &  &  &  & \\
Grasslands &  & \textbf{89.08} & 89.25 & 92.47 & 96.60 & 91.92 & 93.65 & 88.95 & \textbf{91.17} & \textbf{92.89} & \textbf{96.76} & \textbf{91.99} & \textbf{95.73}\\
Croplands &  & 70.97 & \textbf{86.61} &  &  &  &  & \textbf{75.18} & 81.10 &  &  &  & \\
Grass. veg. aq. or flooded &  & \textbf{88.29} & 85.83 & \textbf{88.29} & 85.83 &  &  & 84.86 & \textbf{95.00} & 86.63 & \textbf{93.61} &  & \\
Bare areas &  & \textbf{71.25} & 62.64 & \textbf{88.03} & 72.54 & \textbf{88.03} & 72.54 & 66.30 & \textbf{67.03} & 84.33 & \textbf{79.58} & 82.86 & \textbf{81.69}\\
Built-up &  & \textbf{94.59} & 68.63 &  &  &  &  & 85.71 & \textbf{70.59} &  &  &  & \\
Open Water seasonal &  & \textbf{80.00} & 27.91 & 85.41 & \textbf{85.71} & 85.41 & \textbf{85.71} & 76.47 & \textbf{30.23} & \textbf{97.41} & 80.71 & \textbf{95.00} & 81.43\\
Open Water permanent &  & 83.46 & \textbf{93.67} &  &  &  &  & \textbf{97.09} & 84.39 &  &  &  & \\
Average &  & \textbf{82.64} & 75.78 & 90.40 & 87.83 & 90.79 & 87.59 & 82.17 & \textbf{75.95} & \textbf{91.89} & \textbf{89.77} & \textbf{92.01} & \textbf{89.30}\\
Overall Accuracy &  & 84.32 &  & 92.44 &  & 92.85 &  & \textbf{84.78} &  & \textbf{93.50} &  & \textbf{93.83} & 
\end{tblr}
}
\end{table}

We performed two different \cred{experiments} for the training of the architectures on the ``21KUQ" tile, and the fine-tuning on the ``21KXT" tile. In the first experiment, we tested the effectiveness of the penalty term during the training phase, validating \cred{results on the} photointerpreted samples and testing on 70\% of the \ac{HRLC} samples. Table \ref{tab:training} shows the comparison in terms of \ac{UA} and \ac{PA} accuracies obtained on the photointerpreted ground truth from the reference tile, \cred{for} the Std. \ac{ResNet} and the Hierarchical Class-driven \ac{ResNet} at the three hierarchical levels. The accuracies for the Std. \ac{ResNet} at the intermediate and macro class levels are obtained aggregating the classified samples in the respective classes at the coarse levels. Focusing on the micro class classification, from the table one can see that the proposed methodology obtains higher \ac{PA}s with respect to the Std. \ac{ResNet}, which instead shows better \ac{UA}s. Nonetheless, the proposed methodology performs better compared to the Std. \ac{ResNet} when comparing the \ac{UA}s and \ac{PA}s at the intermediate and macro levels, showing that the network is learning the features hierarchically and penalizing the classification of the targets that do not respect the hierarchy. 

\begin{figure}
    \centering
    \subfigure[]{\includegraphics[width=0.24\textwidth]{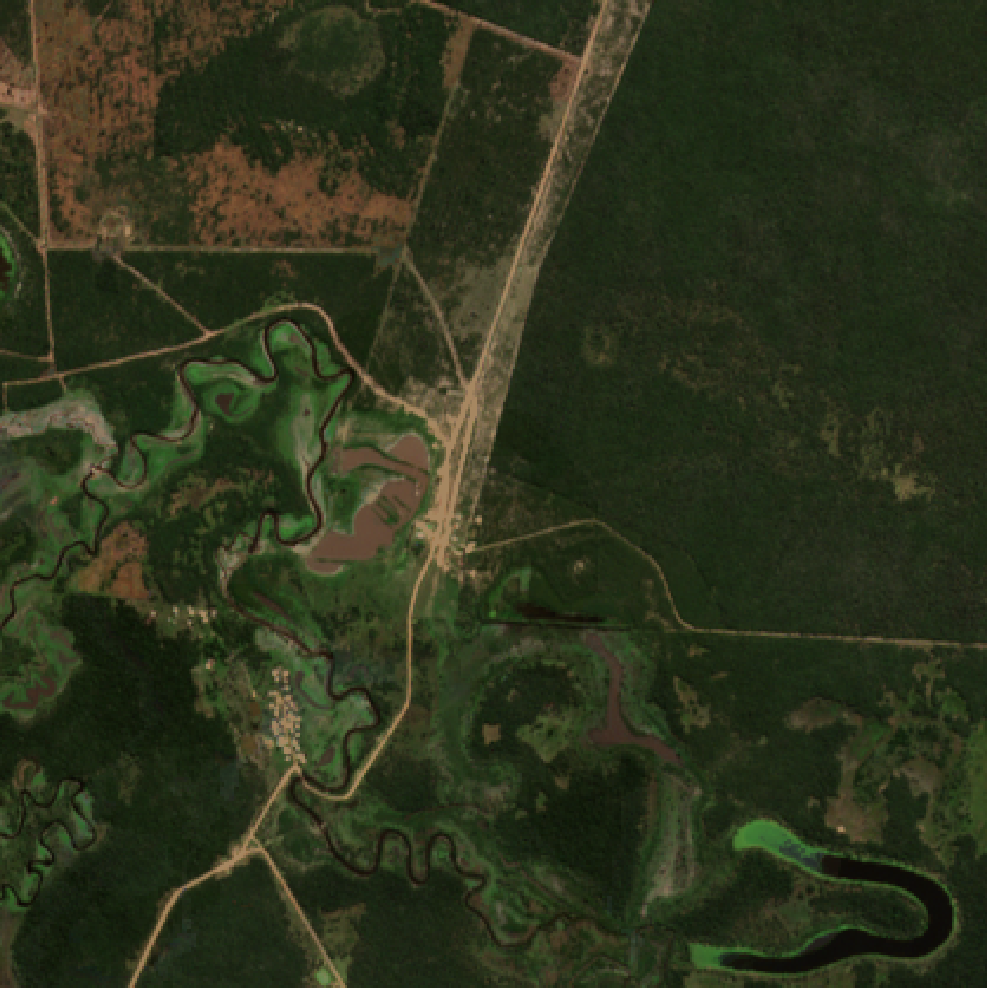}} 
    \subfigure[]{\includegraphics[width=0.24\textwidth]{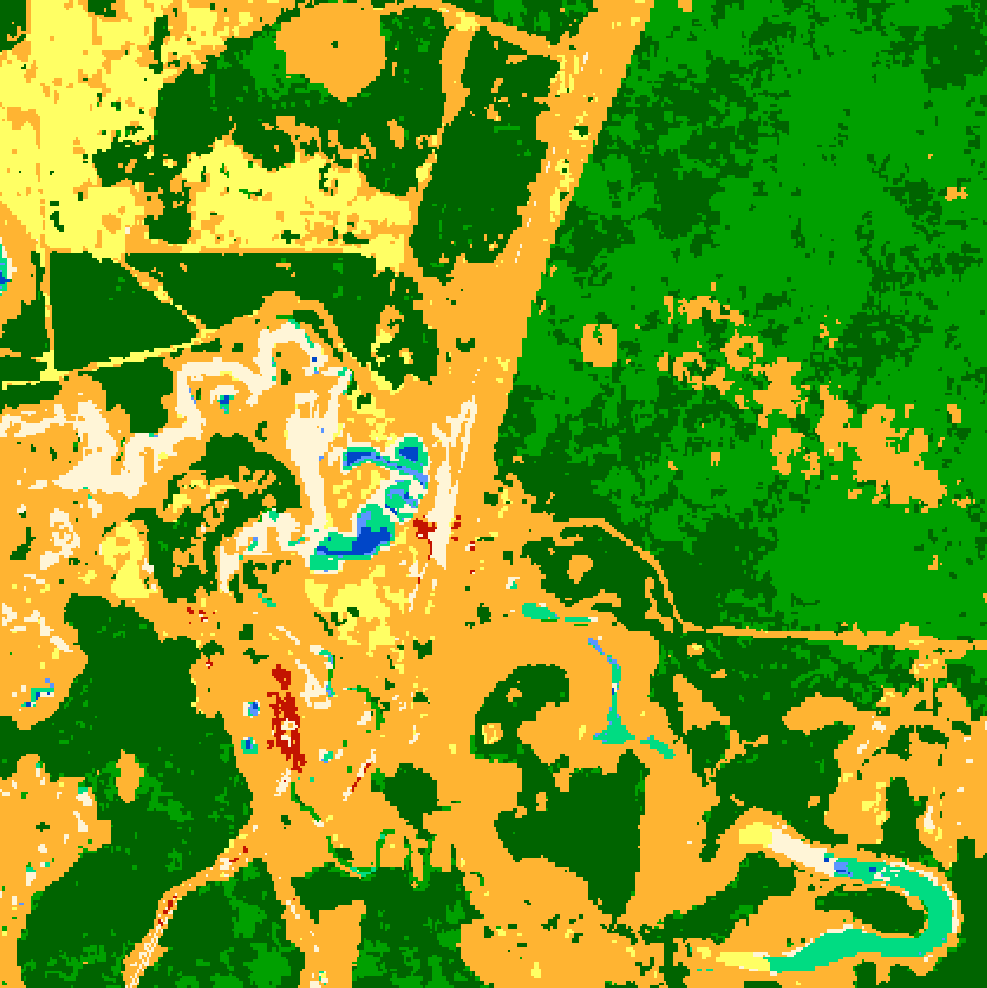}} 
    \subfigure[]{\includegraphics[width=0.24\textwidth]{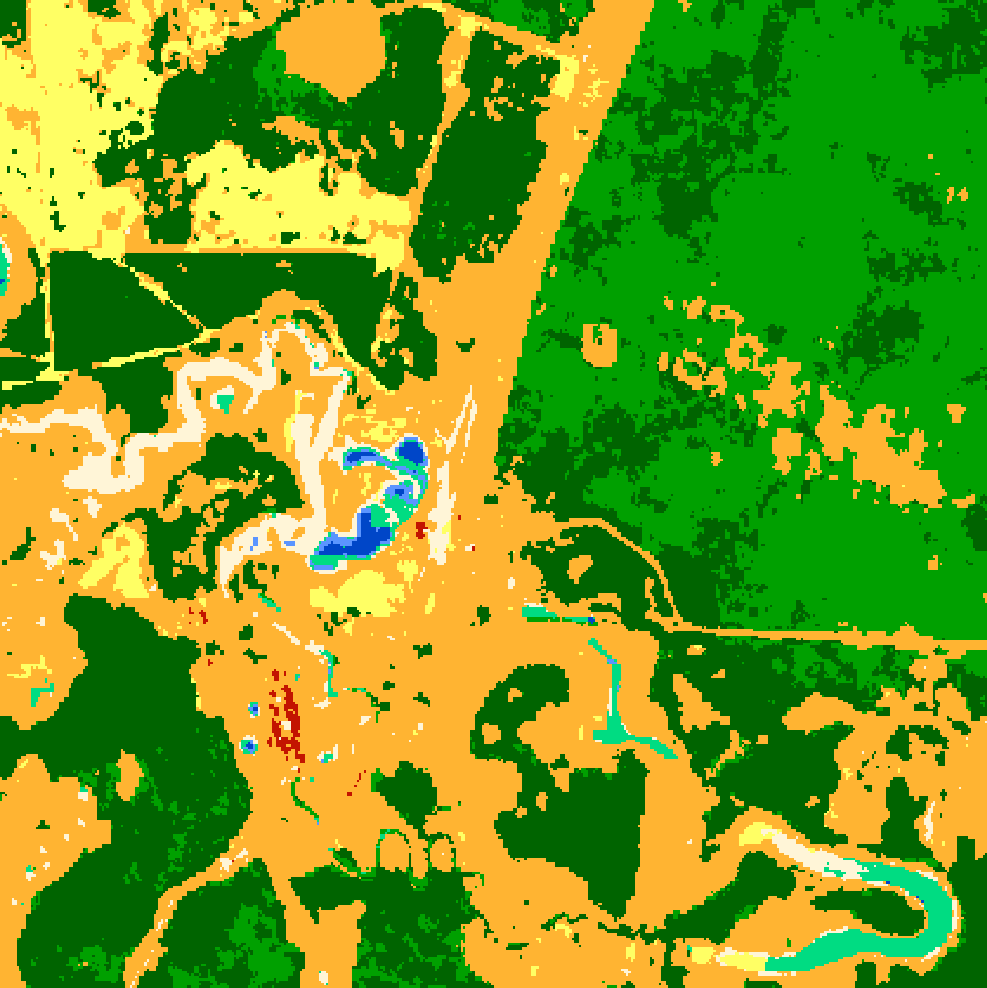}}
    \subfigure[]{\includegraphics[width=0.24\textwidth]{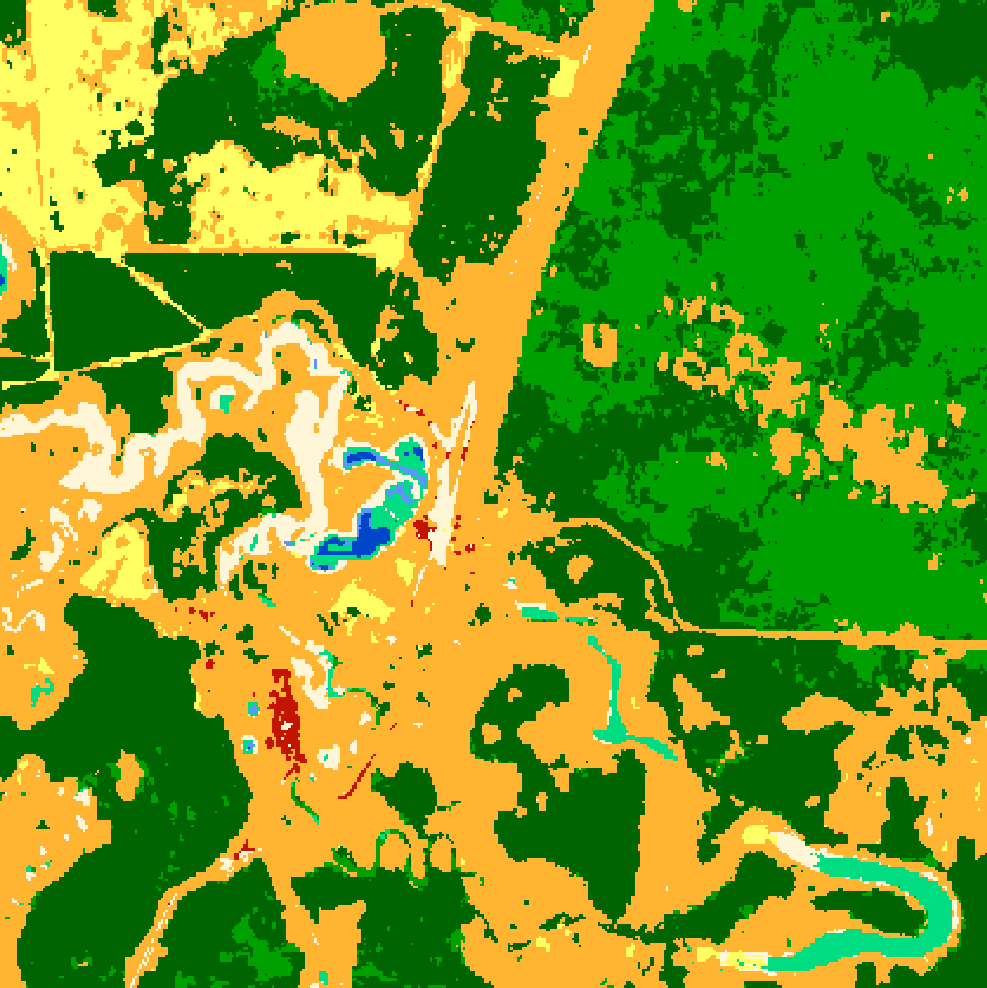}}
    \caption{Qualitative example of the classification \cred{maps} at the micro-class level obtained on a portion of the ``21KUQ" tile: (a) reference monthly composite of July 2019, (b) reference HRLC test set, (c) classification obtained using the Std. \ac{ResNet} and (d) classification \cred{obtained} considering the proposed hierarchical class-driven \ac{ResNet}.}
    \label{fig:qualitative}
\end{figure}

Table \ref{tab:full-comparison-training} shows a comparison of the \ac{F1}-scores calculated considering the photointerpreted ground truth and the \ac{HRLC} labels. The Class-driven \ac{ResNet} outperforms the Std. \ac{ResNet}, showing better F-Scores at the three hierarchical levels considered. In particular, the Class-driven \ac{ResNet} shows better F-Scores in the minoritarian classes, namely ``Grassland Vegetation Aquatic or Regularly Flooded", ``Bare Areas", ``Open Water Seasonal", and ``Open Water Permanent". Focusing on the classification at the intermediate and macro class levels, the hierarchical \ac{ResNet} shows better F-Scores compared to its counterpart. Figure \ref{fig:qualitative} shows a qualitative example of the classification \cred{maps generated} by the Std. \ac{ResNet} and the Class-driven Hierarchical \ac{ResNet}. From the example, one can see that the minoritarian classes, such as ``Built-up" and ``Bare Areas" are better represented by the proposed methodology.

\begin{table}[h!]
\caption{Comparison of the best \ac{F1} scores obtained \cred{both} on the photointerpreted ground truth and the test set extracted from the \ac{HRLC} in the tile ``21KUQ" at the three hierarchical aggregations of the classes adopted, considering the Std. \ac{ResNet} and the proposed Hierarchical Class-driven \ac{ResNet}.}
\label{tab:full-comparison-training}
\centering
\resizebox{\textwidth}{!}{
\begin{tblr}{
  column{3} = {c},
  column{4} = {c},
  column{5} = {c},
  column{6} = {c},
  column{7} = {c},
  column{8} = {c},
  column{9} = {c},
  column{10} = {c},
  column{11} = {c},
  column{12} = {c},
  column{13} = {c},
  column{14} = {c},
  cell{1}{3} = {c=6}{},
  cell{1}{9} = {c=6}{},
  cell{2}{3} = {c=3}{},
  cell{2}{6} = {c=3}{},
  cell{2}{9} = {c=3}{},
  cell{2}{12} = {c=3}{},
  cell{4}{1} = {r},
  cell{4}{2} = {TreeEvergreen},
  cell{4}{3} = {Alto},
  cell{4}{4} = {r=2}{},
  cell{4}{5} = {r=2}{},
  cell{4}{7} = {r=2}{Alto},
  cell{4}{8} = {r=2}{Alto},
  cell{4}{9} = {Alto},
  cell{4}{10} = {r=2}{},
  cell{4}{11} = {r=2}{},
  cell{4}{13} = {r=2}{Alto},
  cell{4}{14} = {r=2}{Alto},
  cell{5}{1} = {r},
  cell{5}{2} = {TreeDeciduous},
  cell{5}{3} = {Alto},
  cell{5}{9} = {Alto},
  cell{6}{1} = {r},
  cell{6}{2} = {Grasslands},
  cell{6}{4} = {r=2}{},
  cell{6}{5} = {r=3}{},
  cell{6}{6} = {Alto},
  cell{6}{7} = {r=2}{Alto},
  cell{6}{8} = {r=3}{Alto},
  cell{6}{10} = {r=2}{},
  cell{6}{11} = {r=3}{},
  cell{6}{12} = {Alto},
  cell{6}{13} = {r=2}{Alto},
  cell{6}{14} = {r=3}{Alto},
  cell{7}{1} = {r},
  cell{7}{2} = {Croplands},
  cell{7}{6} = {Alto},
  cell{7}{12} = {Alto},
  cell{8}{1} = {r},
  cell{8}{2} = {GrassVegetation},
  cell{8}{6} = {Alto},
  cell{8}{7} = {Alto},
  cell{8}{12} = {Alto},
  cell{8}{13} = {Alto},
  cell{9}{1} = {r},
  cell{9}{2} = {BareAreas},
  cell{9}{3} = {Alto},
  cell{9}{4} = {r=2}{},
  cell{9}{5} = {r=2}{},
  cell{9}{6} = {Alto},
  cell{9}{7} = {r=2}{Alto},
  cell{9}{8} = {r=2}{Alto},
  cell{9}{10} = {r=2}{},
  cell{9}{11} = {r=2}{},
  cell{9}{12} = {Alto},
  cell{9}{13} = {r=2}{Alto},
  cell{9}{14} = {r=2}{Alto},
  cell{10}{1} = {r},
  cell{10}{2} = {Builtup},
  cell{10}{3} = {Alto},
  cell{10}{12} = {Alto},
  cell{11}{1} = {r},
  cell{11}{2} = {WaterSeasonal},
  cell{11}{4} = {r=2}{},
  cell{11}{5} = {r=2}{},
  cell{11}{6} = {Alto},
  cell{11}{7} = {r=2}{Alto},
  cell{11}{8} = {r=2}{Alto},
  cell{11}{10} = {r=2}{},
  cell{11}{11} = {r=2}{},
  cell{11}{12} = {Alto},
  cell{11}{13} = {r=2}{Alto},
  cell{11}{14} = {r=2}{Alto},
  cell{12}{1} = {r},
  cell{12}{2} = {WaterPermanent},
  cell{12}{6} = {Alto},
  cell{12}{12} = {Alto},
  cell{13}{1} = {r},
  cell{13}{6} = {Alto},
  cell{13}{7} = {Alto},
  cell{13}{8} = {Alto},
  cell{13}{12} = {Alto},
  cell{13}{13} = {Alto},
  cell{13}{14} = {Alto},
  cell{14}{1} = {r},
  cell{14}{6} = {Alto},
  cell{14}{7} = {Alto},
  cell{14}{8} = {Alto},
  cell{14}{9} = {Alto},
  cell{14}{13} = {Alto},
  cell{14}{14} = {Alto},
  vline{9} = {1,2}{},
  vline{4,7,10} = {2}{},
  vline{6,9,12} = {3,13-14}{},
  vline{2-3,6,9,12} = {4,5,6,7,8,9,10,11,12}{},
  vline{2-3} = {5,7-8,10,12}{},
  hline{3,14} = {3-14}{},
  hline{4} = {2-14}{},
  hline{5-12} = {2}{},
  hline{13} = {-}{},
}
 &  & Photointerpreted GT &  &  &  &  &  & Test set &  &  &  &  & \\
 &  & Std. ResNet &  &  & Class-driven ResNet &  &  & Std. ResNet &  &  & Class-driven ResNet &  & \\
 &  & {Micro\\F1 (\%)} & {Int.\\F1 (\%)} & {Macro\\F1 (\%)} & {Micro\\F1 (\%)} & {Int.\\F1 (\%)} & {Macro\\F1 (\%)} & {Micro\\F1 (\%)} & {Int.\\F1 (\%)} & {Macro\\F1 (\%)} & {Micro\\F1 (\%)} & {Int.\\F1 (\%)} & {Macro\\F1 (\%)}\\
Tree cover
  evergreen &  & \textbf{84.46} & 98.14 & 98.14 & 84.21 & \textbf{98.21} & \textbf{98.28} & \textbf{82.95} & 96.80 & 96.80 & 82.89 & \textbf{97.03 } & \textbf{96.91}\\
Tree cover deciduous &  & \textbf{82.28} &  &  & 80.18 &  &  & \textbf{85.00} &  &  & 83.36 &  & \\
Grasslands &  & 89.17 & 94.49 & 92.78 & \textbf{90.05} & \textbf{94.78} & \textbf{93.83} & 92.79 & 96.45 & 96.45 & \textbf{93.63} & \textbf{96.71} & \textbf{96.57}\\
Croplands &  & 78.01 &  &  & \textbf{78.03} &  &  & 72.75 &  &  & \textbf{72.90} &  & \\
Grass. veg. aq. or flooded &  & 87.04 & 87.04 &  & \textbf{89.65} & \textbf{89.99} &  & 82.88 & 82.88 &  & \textbf{84.80} & \textbf{84.98} & \\
Bare areas &  & \textbf{66.67} & 79.54 & 79.54 & \textbf{66.67} & \textbf{81.88} & \textbf{82.27} & 53.87 & 56.45 & 56.45 & \textbf{61.51} & \textbf{63.96} & \textbf{62.51}\\
Built-up &  & \textbf{79.55} &  &  & 77.42 &  &  & 51.09 &  &  & \textbf{52.77} &  & \\
Open Water seasonal &  & 41.38 & 85.56 & 85.56 & \textbf{43.33} & \textbf{88.28} & \textbf{87.69} & 31.05 & 87.17 & 87.17 & \textbf{39.22} & \textbf{88.21} & \textbf{87.91}\\
Open Water permanent &  & 88.27 &  &  & \textbf{90.29} &  &  & 87.78 &  &  & \textbf{89.42} &  & \\
Average &  & 77.42 & 88.96 & 89.00 & \textbf{77.76} & \textbf{90.63} & \textbf{90.52} & 71.13 & 83.95 & 84.22 & \textbf{73.39} & \textbf{86.18} & \textbf{85.98}\\
Weighted Average &  & 84.00 & 92.33 & 92.77 & \textbf{84.49} & \textbf{93.44} & \textbf{93.76} & \textbf{87.22} & 96.48 & 96.53 & 87.14 & \textbf{96.75} & \textbf{96.65}
\end{tblr}
}
\end{table}

The second experiment consists in the fine-tuning of the pre-trained architecture during the first step on the new tile ``21KXT", considering the addition of the new micro-class ``Shrubs". Table \ref{tab:fine-tuning} shows the comparison between the pre-trained Std. \ac{ResNet} and the Class-driven Hierarchical \ac{ResNet} pre-trained in the previous step. From the table one can see that overall the two networks obtain similar results, with an overall accuracy of 89.13\% obtained by the Std. \ac{ResNet} against 88.12\% of the proposed method. However, the Class-driven \ac{ResNet} obtains higher average \ac{UA}\% and \ac{PA}\% compared to its counterpart, showing that it is \cred{more accurate in} classifying the minoritarian classes. \cred{The hierarchical pre-trained architecture retains its capability to accurately represent the minoritarian classes during the fine-tuning, even when the class has \cblue{less} than 50 samples. Notably, both the architectures were initially pre-trained considering a balanced oversampling of the original training set. However, the hierarchical architecture appears to preserve this balance even \cblue{when} the fine-tuning phase progresses.} For the sake of completeness, we also report the best model trained from scratch on the ground truth. As expected, the \ac{ResNet} trained from scratch performs worse than the two pre-trained network, overfitting the training set. As mentioned earlier, the insufficient amount of training samples poses a challenge in effectively training the architecture from scratch. This results in an unstable classification, which fails to accurately represent the defined classes. 

\begin{table}
\centering
\caption{Comparison of the best \ac{UA} and \ac{PA} accuracies obtained on the test set of the HRLC in the tile ``21KXT" considering the Std. \ac{ResNet} and the proposed Hierarchical Class-driven \ac{ResNet}. The table also reports the performance of the architecture trained from scratch.}
\label{tab:fine-tuning}
\begin{tblr}{
  column{3} = {c},
  column{4} = {c},
  column{5} = {c},
  column{6} = {c},
  column{7} = {c},
  column{8} = {c},
  cell{1}{3} = {c=6}{},
  cell{2}{3} = {c=2}{},
  cell{2}{5} = {c=2}{},
  cell{2}{7} = {c=2}{},
  cell{4}{1} = {r},
  cell{4}{2} = {TreeEvergreen},
  cell{4}{5} = {Alto},
  cell{4}{8} = {Alto},
  cell{5}{1} = {r},
  cell{5}{2} = {TreeDeciduous},
  cell{5}{6} = {Alto},
  cell{5}{7} = {Alto},
  cell{6}{1} = {r},
  cell{6}{2} = {Shrubs},
  cell{6}{5} = {Alto},
  cell{6}{8} = {Alto},
  cell{7}{1} = {r},
  cell{7}{2} = {Grasslands},
  cell{7}{6} = {Alto},
  cell{7}{7} = {Alto},
  cell{8}{1} = {r},
  cell{8}{2} = {Croplands},
  cell{8}{7} = {Alto},
  cell{8}{8} = {Alto},
  cell{9}{1} = {r},
  cell{9}{2} = {GrassVegetation},
  cell{9}{5} = {Alto},
  cell{9}{8} = {Alto},
  cell{10}{1} = {r},
  cell{10}{2} = {BareAreas},
  cell{10}{7} = {Alto},
  cell{10}{8} = {Alto},
  cell{11}{1} = {r},
  cell{11}{2} = {Builtup},
  cell{11}{6} = {Alto},
  cell{11}{7} = {Alto},
  cell{12}{1} = {r},
  cell{12}{2} = {WaterSeasonal},
  cell{12}{6} = {Alto},
  cell{12}{7} = {Alto},
  cell{13}{1} = {r},
  cell{13}{2} = {WaterPermanent},
  cell{13}{5} = {Alto},
  cell{13}{6} = {Alto},
  cell{14}{1} = {r},
  cell{14}{7} = {Alto},
  cell{14}{8} = {Alto},
  cell{15}{1} = {r},
  cell{15}{3} = {c=2}{},
  cell{15}{5} = {c=2}{Alto},
  cell{15}{7} = {c=2}{},
  vline{4,6,8} = {2}{},
  vline{5,7} = {3,14}{},
  vline{2-3,5,7} = {4-13}{},
  vline{4,6} = {15}{},
  hline{3,15} = {3-8}{},
  hline{4} = {2-8}{},
  hline{5-13} = {2}{},
  hline{14} = {-}{},
}
 &  & FT test set &  &  &  &  & \\
 &  & {Std. ResNet\\from scratch} &  & Std. ResNet &  & {Class-driven~\\ResNet} & \\
 &  & {Micro\\UA(\%)} & {Micro\\PA(\%)} & {Micro\\UA(\%)} & {Micro\\PA(\%)} & {Micro\\UA(\%)} & {Micro\\PA(\%)}\\
Tree cover  evergreen & \hphantom{-} & 41.32 & 96.69 & \textbf{84.28} & 84.56 & 79.34 & \textbf{86.33}\\
Tree cover deciduous &  & 11.11 & 0.00 & 79.44 & \textbf{80.55} & \textbf{83.12} & 73.49\\
Shrubs &  & 16.41 & 13.00 & \textbf{34.19} & 29.51 & 26.56 & \textbf{50.44}\\
Grasslands &  & 94.66 & 87.34 & 94.92 & \textbf{95.83} & \textbf{95.76} & 94.93\\
Croplands &  & 1.54 & 3.46 & 71.37 & 57.10 & \textbf{74.59} & \textbf{57.27}\\
Grass. veg. aq. or flooded &  & 46.35 & 71.46 & \textbf{70.04} & 63.53 & 54.99 & \textbf{73.11}\\
Bare areas &  & 9.03 & 20.16 & 44.55 & 33.69 & \textbf{44.78} & \textbf{42.79}\\
Built-up &  & 31.86 & 32.87 & 49.65 & \textbf{55.38} & \textbf{60.83} & 51.83\\
Open Water seasonal &  & 12.07 & 13.57 & 43.53 & \textbf{47.12} & \textbf{57.50} & 30.85\\
Open Water permanent &  & 50.88 & 51.07 & \textbf{76.12} & \textbf{76.13} & 73.92 & 75.70\\
Average &  & 31.52 & 38.96 & 64.81 & 62.34 & \textbf{65.14} & \textbf{63.67}\\
Overall Accuracy &  &  70.38 &  & \textbf{89.13} &  & 88.12 & 
\end{tblr}
\end{table}

\section{Conclusion}
\label{sec:conclusion}
In this paper, we \cred{have} introduced a versatile modular architecture capable of learning features across distinct levels of hierarchy \cred{of semantic representation of land-cover classes}. This architecture can be \cred{exploited and} customized for novel datasets and problem formulations through fine-tuning. The method involves two key aspects: (i) a modification of the standard \ac{ResNet}, incorporating additional classification branches at three hierarchical \cred{class} levels, and (ii) the introduction of penalty terms within the network hierarchy classification to discourage inconsistent hierarchical classifications. \cred{In this study, we defined a hierarchical architecture able to exploit within the same architecture different hierarchical levels of feature extraction. From the \cblue{experimental} results, the class-driven Hierarchical \ac{ResNet} shows an increase of performance, leveraging the hierarchical structure to correctly map the classes within the three hierarchical levels. The proposed network was then successfully used as a backbone for a new classification task, allowing the characterization of new tasks and challenges even with a limited set of labelled samples. The results reported on the two amazonian tiles \cblue{of Sentinel-2} analysed demonstrate the effectiveness of the proposed method.  Moreover,} the proposed \cred{architecture} is promising for \cred{modelling} complex classification schemes, overall showing better performance \cred{on the} minoritarian classes. Although the proposed network proved its capability to mitigate highly imbalanced classification problem, we aim to further investigate viable solutions to better handle this issue. Furthermore, we demonstrated the feasibility of fine-tuning the proposed network in \ac{TS} with a diverse geographical context, characterized by different properties, and incorporate new classes in the hierarchy. Our results in the fine-tuning indicate a marginal performance increase compared to the training of the network. Nonetheless, the results obtained show promising potential, prompting us to further demonstrate its effectiveness with more detailed datasets and intricate classification schemes in our future works.

\acknowledgments    
\cred{This work was partially supported by the High Resolution Land Cover project, funded by the European Space Agency (ESA), within the Climate Change Initiative (CCI) Programme. }
 
\bibliography{report} 
\bibliographystyle{spiebib} 

\end{document}